%% file: main.tex
\documentclass{article}
\usepackage{multirow}
\usepackage[preprint]{corl_2026} 
\usepackage{amsmath,amssymb}
\usepackage{microtype}
\usepackage{placeins}
\usepackage{needspace}
\usepackage{float}
\usepackage{graphicx}
\usepackage{algorithm}
\usepackage{algpseudocode}
\usepackage{capt-of}
\usepackage{enumitem}

\usepackage{booktabs}       
\usepackage{amsfonts}       
\usepackage{nicefrac}       

\usepackage{mathrsfs}
\usepackage{mathtools}
\usepackage{amssymb}
\usepackage{enumitem}

\usepackage{wrapfig}
\usepackage{physics}
\usepackage{overpic}
\usepackage{subcaption}

\title{\fontsize{16pt}{19pt}\selectfont Beyond Semantic Classes: Online Physical Parameter Estimation via Embedding-Conditioned Gaussian Mixtures}

\author{
  Alexander Borca-Tasciuc \quad Prayag Sharma \quad Franck Djeumou\\
  Rensselaer Polytechnic Institute, USA \quad \texttt{\{borcaa,sharmp6,djeumf2\}@rpi.edu}
}

\begin{document}
\maketitle
\raggedbottom

\vspace*{-5mm}
\begin{abstract}
Enabling robots to infer interaction-relevant physical properties from vision before contact would improve planning and control in unstructured environments. Vision provides useful pre-contact information, but appearance alone is often ambiguous: visually similar regions can induce different friction, stiffness, or sinkage responses upon contact. Existing approaches often collapse this ambiguity to a point estimate or aggregate experience through fixed semantic classes, limiting their ability to preserve distinct physical regimes. Instead, we propose a real-time estimator that replaces semantic class indexing with an interaction-grounded Gaussian-mixture posterior over latent physical parameters, conditioned directly on continuous visual embeddings. Our approach maintains interaction-derived Gaussian parameter estimates, typically from inverse modeling, as mixture components while using embedding similarity to retrieve and weight relevant past interactions for each new query region. A support-aware fusion mechanism incorporates a conditional generative prior model when interaction history is sparse, while a learned embedding refinement improves retrieval relevance without discretizing the representation. Our experiments show that the estimator recovers multimodal physical beliefs, rejects unrelated interaction history, avoids spatial map discretization, and evaluates at controller rates. In a high-fidelity autonomous racing simulation with unknown low- and high-friction surface regions, the posterior enables the controller to adapt over repeated laps as interaction evidence accumulates, improving performance while maintaining friction-aware behavior.
\end{abstract}

\keywords{Parameter Estimation, Multimodal Beliefs, Vision-Based Control}
\vspace*{-2mm}
\section{Introduction}


Autonomous systems operating in unstructured environments select control actions whose outcomes depend on physical parameters of the surface ahead
that cannot be measured at range. Model-based controllers that reason over uncertainty in these parameters can bound risk against adverse outcomes, but doing so requires a distributional estimate rather than a point prediction. Vision provides pre-contact evidence, yet visual appearance does not uniquely determine physical parameters: dry asphalt and black ice can appear similar while differing in friction by an order of magnitude.
A pre-contact estimate should therefore preserve the set of plausible parameter values rather than collapse them to a single prediction. 
Since vision cannot resolve this ambiguity on its own, interaction measurements provide the missing physical evidence: they reveal which parameter regime was realized and should reshape the belief assigned to visually similar regions encountered later. This paper investigates how to maintain a real-time, interaction-based multimodal belief over latent physical parameters by combining visual embeddings with post-contact measurements, producing a posterior that can inform downstream planning and control under physical-parameter uncertainty.


\begin{figure}[!hbt]
    \centering
    \vspace*{-2mm}
    \includegraphics[width=1.0\linewidth]{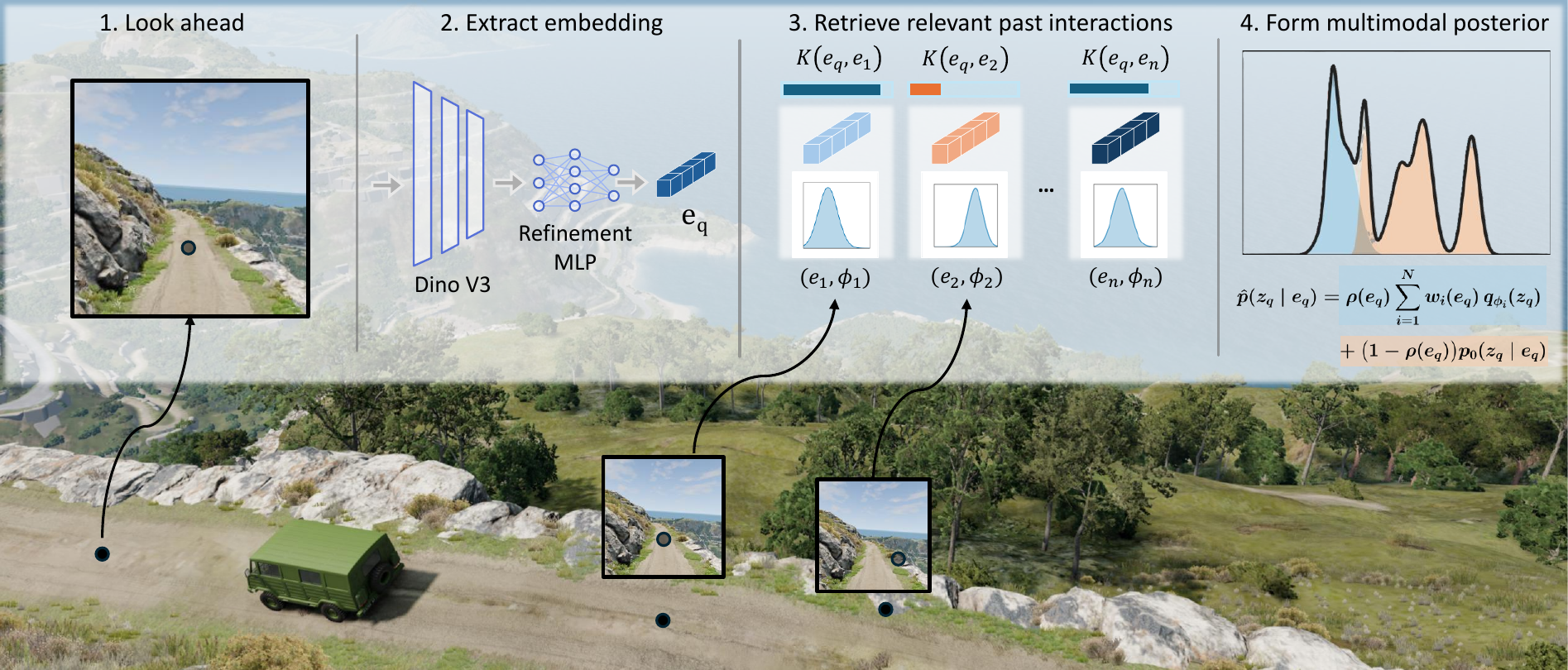}
    \vspace*{-4mm}
    \caption{Our approach leverages visual embeddings to retrieve interaction-derived Gaussian beliefs, which are fused with a prior to produce a multimodal posterior for planning and control.}
    \vspace*{-5mm}
    \label{fig:main_figure}
\end{figure}
\vspace{-0.5mm}
Existing vision-based physical prediction methods \citep{Sabanovic2020RoadSurface, Ojala2024WinterFriction, Gasparino2022WayFAST, Margolis2023ActiveSensing}, use appearance to estimate terrain parameters before contact, but they typically return a single nominal value. This discards information when the same visual evidence is consistent with multiple physical regimes. Methods that incorporate interaction measurements address a different part of the problem: they can revise parameter estimates after contact and propagate those updates to future predictions. However, existing formulations typically organize this experience through a fixed set of semantic terrain or material classes \citep{Ewen2024FeelIt,Greiff2026Dynamic}. In these methods, the semantic label becomes the unit of generalization: measurements from one region update a class-level Gaussian that is reused for other regions with the same label. This creates a representation bottleneck. Physically distinct surfaces that share a label are averaged into a single unimodal estimate, while materials outside the vocabulary must be forced into an existing class, shifting that class's estimate toward an unrelated parameter regime. The resulting belief is constrained by the semantic vocabulary rather than the continuous structure of visual observations.


We propose to replace semantic class indexing with a posterior conditioned on continuous visual embeddings. Each interaction produces a Gaussian estimate of the latent physical parameter through inverse modeling, which we store together with the visual embedding of the corresponding region. For a new query, a similarity kernel over embeddings assigns weights to estimates from visually related past interactions, forming a Gaussian mixture posterior. Because the interaction-derived estimates remain separate mixture components, the posterior can represent multiple physical regimes for visually similar terrain rather than averaging them into a single class-level Gaussian. When the query is weakly supported by prior interactions, we blend the history-based mixture with a conditional prior over Gaussian parameter estimates; when instantiated as a diffusion model, samples from this prior contribute additional Gaussian components while keeping the final posterior explicit, efficient to evaluate or sample on the GPU, and directly usable inside a real-time model-based controller.


In practice, our formulation depends on the geometry of the visual embedding space. Similarity in that space should reflect relevance for physical-parameter prediction, but raw pretrained embeddings are also sensitive to nuisance factors such as lighting, viewpoint, camera pose, and residual positional structure. These effects can separate observations that should share interaction evidence, or make irrelevant observations appear close under the kernel. We therefore learn an input-dependent refinement of the visual embeddings before computing similarity. The refinement improves retrieval of relevant interaction measurements while avoiding a fixed semantic vocabulary.

Together, these pieces turn visual prediction into an interaction-based posterior that remains multimodal, tractable, and fast enough for model-based control. Our contributions are as follows:
\begin{itemize}[itemsep=-1mm, topsep=-1mm]
    \item An interaction-grounded Gaussian-mixture posterior over latent physical parameters, conditioned on continuous visual embeddings, that preserves 
    multimodal variation without semantic class indexing.

    \item A support-aware fusion mechanism that combines interaction history with a generative prior; when the prior is defined to output Gaussian belief parameters, the posterior remains an explicit mixture that is efficient to evaluate and sample.

    \item A retrieval-space refinement that improves embedding similarity without discretizing the visual representation, together with a real-time evaluation in a closed-loop, high-fidelity autonomous racing simulation with unknown low- and high-friction surface regions.

\end{itemize}

\section{Related Work}
\vspace*{-2mm}
Visual observations are widely used to infer terrain properties before contact. Early work used appearance and material recognition to estimate terrain friction, demonstrating that pre-contact visual evidence can inform physical interaction \citep{Brandao2017MaterialFriction}. Subsequent methods predict traction, traversability, speed, cost, or dynamics from visual input for off-road planning and control \citep{Gasparino2022WayFAST,Mattamala2024WVN,Triest2025Velociraptor,Gibson2024Dynamics}, including recent approaches that use simulation-derived labels \citep{Chen2024Physical}, cross-modal supervision \citep{Margolis2023ActiveSensing}, or physics decoders \citep{Agishev2024MonoForce} to predict quantities such as friction and stiffness. These methods establish vision as a useful source of pre-contact physical information, but most do not return a distribution over the latent physical parameters.
%
Some other approaches explicitly represent uncertainty in the predicted quantity. For example, learned speed-distribution maps estimate distributions over achievable speed for risk-aware planning \citep{Triest2022SpeedMap}. However, such methods do not update the physical-parameter belief using interaction measurements collected after contact.


Prior work has also used tactile, proprioceptive, or other measurements collected through interaction to revise physical-parameter estimates after contact. Semantic-property mapping methods associate terrain or material classes with probabilistic estimates of physical quantities, allowing measurements collected at one location to inform predictions elsewhere through the shared class representation. \citep{Ewen2022Maps} maintain probabilistic terrain-property estimates indexed by semantic class, and \citep{Ewen2024FeelIt} jointly update semantic predictions and physical-property estimates from visual and tactile measurements. Related filtering approaches model evolving associations between semantic classes and inferred dynamical parameters, allowing the estimated physical properties of a class to change as new measurements are observed \citep{Greiff2026Dynamic}. These methods provide a principled way to combine visual prediction with physical evidence collected through interaction, though the fixed semantic representation can cause class-level beliefs to average distinct physical regimes within a label and be corrupted when out-of-vocabulary materials are absorbed into an existing class.

Large-scale visual pretraining has produced continuous embeddings that transfer across robotic perception, mapping, navigation, and prediction tasks. Vision foundation models \citep{Radford2021CLIP,Simeoni2025DINOv3} such as CLIP and DINO provide high-dimensional features used for traversability learning, off-road navigation, open-dictionary mapping, and dynamics prediction \citep{Mattamala2024WVN,Triest2025Velociraptor,Wilson2025LatentBKI,Gibson2024Dynamics}. These works show that continuous representations can support robotic prediction without reducing every observation to a fixed semantic label. However, pretrained embeddings are not guaranteed to provide the right similarity geometry for physical-parameter prediction: nuisance factors such as lighting, distance, viewpoint, occlusion, or residual positional structure can affect nearest-neighbor relationships in feature space. We therefore refine the embedding space so that similarity better reflects relevance for physical interaction, rather than generic visual or semantic similarity. In contrast to using continuous embeddings only as features for prediction, we use them as the conditioning space for an interaction-grounded posterior over latent physical parameters. For probabilistic prediction from these continuous representations, Gaussian-process regression offers another approach, with sparse approximations that can support real-time prediction~\citep{Titsias2009SparseGP,Schreiter2015RealTimeGP}. However, standard Gaussian-likelihood formulations produce a single Gaussian predictive belief at each query, which cannot preserve distinct physical modes associated with similar visual observations. Their Gaussian priors also restrict the distributional form available when interaction evidence is limited. Sparse inducing-point methods additionally compress accumulated evidence into a finite representation, introducing an approximation whose quality depends on the number and placement of inducing points.


\vspace*{-3mm}
\section{Method}
\label{sec:method}
\vspace*{-3mm}
We consider the problem of estimating a latent physical parameter $z_q \in \mathbb{R}^d$ associated with a query region observed before contact. The parameter may encode friction, stiffness, sinkage, or other local quantities that govern the environment's interaction response. The estimator receives a continuous visual embedding $e_q \in \mathbb{R}^m$ for the query region and an interaction history $\mathcal{D}_N = \{(e_i,\phi_i)\}_{i=1}^{N}$, where $e_i$ is the embedding associated with a previously traversed region and $\phi_i$ parameterizes an uncertain estimate of the corresponding physical parameter. Our goal is to construct an interaction-based predictive belief $\hat p(z_q|e_q; \mathcal{D}_N)$ given $\mathcal{D}_N$, without reducing the visual observation to a discrete semantic class. 
In what follows, we use $ p(\cdot | e_q)$ instead of $p(\cdot | e_q; \mathcal{D}_N)$ when clear from context.

\paragraph{Trajectory-based local parameter estimates.}
Each entry in $\mathcal{D}_N$ summarizes the physical evidence obtained from a short interaction trajectory. Let $\tau_i$ denote the measurements collected while traversing a given region, such as states, controls, slip, contact forces, or motion residuals. Given a dynamics or interaction model, these measurements can be used to fit a local physical parameter and quantify uncertainty in that estimate. We represent the resulting local belief as
\begin{equation}
    q_{\phi_i}(z) := p(z | \tau_i)
    \approx
    \mathcal{N}(z;\mu_i,\Sigma_i),
    \text{ where }
    \phi_i=(\mu_i,\Sigma_i).
\end{equation}
For example, $\mu_i$ may be obtained by minimizing a trajectory-level negative log-likelihood \cite{djeumou2024diff}, with $\Sigma_i$ estimated from a Laplace approximation to the local objective or from the covariance of the upstream estimator \citep{kristiadi2021learnable, kristiadi2023promises}. Our method is agnostic to this local identification procedure: it assumes only that each interaction provides a Gaussian belief over the local physical parameter. The history, therefore, stores uncertain physical hypotheses, rather than point estimates, together with the visual embeddings of the regions where those hypotheses were observed.

\paragraph{Embedding-conditioned history posterior via kernel weighting.}
The history $\mathcal{D}_N$ provides a set of local Gaussian beliefs indexed by continuous visual embeddings. 
For a query embedding $e_q$, we use similarity in this space to transfer interaction evidence from previously traversed regions to the query. Let $K_h(e_q,e_i)\geq 0$ be a kernel measuring the relevance of historical embedding $e_i$ to $e_q$, with bandwidth parameter $h$. In our implementation, embeddings are normalized, and $K_h$ is an exponential kernel of cosine similarity, although the construction only requires a nonnegative kernel.

The kernel induces a normalized relevance weight for each prior interaction, which we use to define a kernel conditional estimate over the local belief parameters stored in the history as follows:
\begin{equation}
    \hat{p}_{\mathrm{hist}}(\phi | e_q)
    =
    \sum\nolimits_{i=1}^{N}
    w_i(e_q)\delta_{\phi_i}(\phi), \text{ where }
    w_i(e_q)
    =
    \frac{K_h(e_q,e_i)}
    {\sum_{j=1}^{N} K_h(e_q,e_j)}.
\end{equation}
Here, $\delta_{\phi_i}$ denotes a point mass at the Gaussian parameters $\phi_i$.
The distribution $\hat p_{\mathrm{hist}}(\phi | e_q)$ can be viewed as a retrieval distribution over interaction-derived physical hypotheses, assigning larger mass to local beliefs whose embeddings are more relevant to the query.
The predictive belief induced by the interaction history is obtained by marginalizing over these local belief parameters:
\begin{equation}
    \hat{p}_{\mathrm{hist}}(z_q | e_q)
    =
    \int q_{\phi}(z_q)\,
    \hat{p}_{\mathrm{hist}}(\phi | e_q)\,d\phi
    =
    \sum_{i=1}^{N}
    w_i(e_q)\,q_{\phi_i}(z_q)
    =
    \sum_{i=1}^{N}
    w_i(e_q)\,
    \mathcal{N}(z_q;\mu_i,\Sigma_i).
    \label{eq:history_posterior}
\end{equation}
Thus, the history posterior is an explicit Gaussian mixture whose components are the uncertain local parameter beliefs obtained from previous interactions. Because each interaction-derived belief remains a separate mixture component, the estimator can preserve multiple physical regimes associated with visually similar observations instead of collapsing them into a single averaged estimate.

\paragraph{Support-aware fusion with a generative prior model.}
The history posterior in Eq.~\eqref{eq:history_posterior} is normalized over the stored interactions. As a result, it defines a valid mixture even when the query embedding is far from all the embeddings in the history. In such cases, the relative weights $w_i(e_q)$ may still select the closest stored beliefs, but the history itself provides little evidence that any of them are physically relevant to the query. We therefore separate relative retrieval within the history from the absolute support of the query under the history.

We measure support using the unnormalized kernel mass $M(e_q)=\sum\nolimits_{i=1}^{N}K_h(e_q,e_i)$ and convert it to a history-confidence coefficient $\rho(e_q)=\frac{M(e_q)}{M(e_q)+\kappa}$, where $\kappa>0$ controls how much embedding-space support is required before relying on interaction history. When $e_q$ is close to many stored embeddings, $M(e_q)$ is large and $\rho(e_q)$ approaches one. When the query is weakly supported, $\rho(e_q)$ decreases and the estimator falls back toward a prior predictive distribution.

The final belief interpolates between the history-induced posterior and a conditional prior,
\begin{equation}
\label{eq:support_fusion}
    \hat p (z_q | e_q)
    =
    \rho(e_q)\hat p_{\mathrm{hist}}(z_q | e_q)
    +
    \bigl(1-\rho(e_q)\bigr)p_0(z_q | e_q).
\end{equation}
This fusion prevents unsupported retrieval from producing overconfident predictions while preserving the interaction-grounded mixture when the query is in a well-sampled region of the embedding space. 

The prior can be specified in either of two forms. The simplest option is a direct physical-parameter prior $p_0^z(z_q | e_q)$, for example a conditional generative model over $z_q$. This prior can be used directly in Eq.~\eqref{eq:support_fusion}; samples from it provide candidate physical parameters when the query has limited support in the history. Alternatively, the prior can be defined over Gaussian belief parameters: $p_0^\phi(\phi | e_q)$ with $\phi=(\mu,\Sigma)$. This form matches the representation used by the interaction history: each prior sample corresponds to a Gaussian belief over the physical parameter rather than a single parameter value. Drawing samples $\phi^{(s)}=(\mu^{(s)},\Sigma^{(s)})$ from a conditional generative model gives the approximation
\begin{equation}
\label{eq:param_prior_samples}
    p_0(z_q | e_q) =
    \int q_{\phi}(z_q)\,p_0^\phi(\phi | e_q)\,d\phi =
    \mathbb{E}_{\phi\sim p_0^\phi(\cdot | e_q)}
    \left[q_{\phi}(z_q)\right]
    \approx
    \frac{1}{S}\sum\nolimits_{s=1}^{S}
    \mathcal{N}(z_q;\mu^{(s)},\Sigma^{(s)}).
\end{equation}
Substituting Eq.~\eqref{eq:param_prior_samples} into
Eq.~\eqref{eq:support_fusion} yields
\begin{equation}
\label{eq:full_mixture}
    \hat p(z_q | e_q)
    =
    \rho(e_q)\sum\nolimits_{i=1}^{N}w_i(e_q)\mathcal{N}(z_q;\mu_i,\Sigma_i)
    +
    \bigl(1-\rho(e_q)\bigr)
    \frac{1}{S}\sum\nolimits_{s=1}^{S}
    \mathcal{N}(z_q;\mu^{(s)},\Sigma^{(s)}).
\end{equation}
Thus, when the prior is generated in Gaussian-belief space, the fused posterior remains an explicit Gaussian mixture whose components come from either interaction history or the prior.

\paragraph{Retrieval-space refinement.}
\label{sec:refinement}
The posterior in Eq.~\eqref{eq:full_mixture} relies on the kernel $K_h(e_q,e_i)$ to identify which interaction-derived beliefs are relevant to a query. This requires an embedding geometry in which similarity reflects physical-parameter relevance, not only generic visual similarity. In practice, raw pretrained embeddings can be sensitive to nuisance variation such as lighting, viewpoint, camera pose, texture bias, or residual positional structure. These effects can distort retrieval: regions with relevant interaction evidence may be separated in embedding space, while physically irrelevant regions may receive high similarity. Appendix~\ref{app:embedding_refinement} provides an empirical illustration of this behavior on our simulated images. We therefore learn a residual refinement of the base visual embedding before
computing kernel similarities. 

Let $\bar e \in \mathrm{R}^m$ denote the normalized embedding
produced by the frozen visual encoder. From an unlabeled image corpus, we cluster
base embeddings using HDBSCAN~\citep{Campello2013HDBSCAN}. HDBSCAN labels embeddings that it does not assign to a cluster as noise. For each non-noise cluster
$k$, we define a prototype direction 
$ c_k = \mathrm{unit}(\sum_{i:a_i=k}\bar e_i)$
where $a_i$ is the cluster assignment and $\mathrm{unit}(v) = v / \|v\|$. These prototypes are used only to train the retrieval embedding; they are not used as terrain classes in the posterior. We train a lightweight residual model $r_\eta$ to move clustered embeddings toward their prototype directions while leaving unsupported samples close to their original embeddings. With $d_i=c_{a_i}-\bar e_i$ for clustered samples, the training objective is
\begin{equation}
\label{eq:retrieval_refinement_loss}
    \mathcal{L}_{\mathrm{ref}}
    =
    \sum\nolimits_{i:a_i\neq \mathrm{noise}}
    \left(
        \left\|r_\eta(\bar e_i)-d_i\right\|_2^2
        +
        \lambda_{\mathrm{cluster}}
        \left\|r_\eta(\bar e_i)\right\|_2^2
    \right)
    +
    \lambda_{\mathrm{noise}}
    \sum\nolimits_{i:a_i=\mathrm{noise}}
    \left\|r_\eta(\bar e_i)\right\|_2^2 .
\end{equation}
The non-noise magnitude penalty regularizes the residual correction, preventing overly large prototype-induced shifts when the HDBSCAN assignments are imperfect or weakly supported. The retrieval embedding used by the kernel is then
$e = \mathrm{unit}(\bar e+\alpha r_\eta(\bar e))$ where $\alpha\in[0,1]$ controls the refinement strength. Setting $\alpha=0$ recovers the frozen embedding, while larger values increase the prototype-based correction before renormalization.

\begin{figure}[!hbt]
    \vspace{-0.5em}
    \includegraphics[width=1\linewidth]{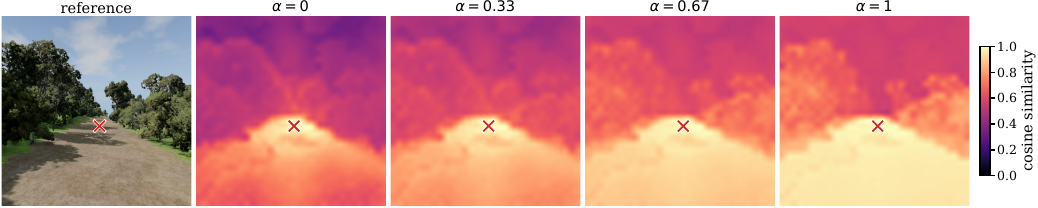}
    \caption{Effect of the refinement strength $\alpha$ on query-induced similarity maps.}
    \vspace{-2mm}
    \label{fig:alpha_sim}
\end{figure}
Fig.~\ref{fig:alpha_sim} visualizes the resulting change in query-induced similarity maps for the refinement model trained on various terrains in our simulated environment, which is used throughout the simulation experiments. Training and experiment details are provided in Appendix~\ref{app:embedding_refinement}. Such a refinement changes only the geometry used for retrieval in $K_h$; it does not discretize the terrain representation, alter the Gaussian components, or replace the continuous embedding-conditioned posterior with a class-based model.

\section{Experiments}
\label{sec:experiments}

We evaluate our approach through four simulation experiments and a real-vehicle parameter-prediction experiment. First, we test whether conditioning on continuous embeddings avoids the class averaging and out-of-vocabulary contamination that arise in baseline approaches with fixed semantic classes. Second, we show that the same posterior construction can use kernels beyond visual similarity by composing the embedding similarity kernel with a spatial proximity kernel. Third, we benchmark history-query latency as the interaction database grows. Fourth, we evaluate the estimator when used for control in a high-fidelity autonomous racing simulation. Finally, we evaluate real-vehicle parameter prediction using recorded trajectories and controls (Section~\ref{sec:hardware}). The simulation experiments are run in BeamNG~\cite{beamngdrive} with a forward-facing camera; images are encoded with DINOv3~\cite{Simeoni2025DINOv3}, refined as in Sec.~\ref{sec:refinement}, and stored with camera poses in a FIFO buffer. Contact points along the trajectory generate uncertain local parameter estimates $\phi_i=(\mu_i,\sigma_i^2)$ from a Normal-Inverse-Gamma model, which are linked to vision by projecting each contact into the buffered frames and adding the closest valid embedding-estimate pair $(e_i,\phi_i)$ to the interaction history. Further details of the simulation setup are given in Appendix~\ref{app:sim_details}.



\vspace*{-2mm}
\subsection{Semantic Class Bottlenecks: Multimodality and Out-of-Vocabulary Contamination}
\vspace*{-2mm}




We compare our estimator with a semantic-property baseline in the style of \cite{Ewen2024FeelIt}. The environment contains grass, dirt, and concrete. We test the estimators in two scenarios: E1 and E2. Grass and dirt have the same unimodal friction distributions in both scenarios. In E1, concrete is included in the semantic vocabulary, and we assign it a bimodal friction distribution representing high-friction dry areas and low-friction areas covered by black ice. This creates two physical regimes within one semantic class, providing a controlled test of whether the predictive belief preserves both modes. In E2, concrete has a unimodal distribution but removed from the semantic vocabulary, so concrete observations are represented by ambiguous semantic evidence over the remaining grass and dirt classes.

\begin{wrapfigure}{r}{0.6\textwidth}
\vspace{-4mm}
\centering
\begin{subfigure}[t]{0.49\linewidth}
\centering
\includegraphics[width=\linewidth]{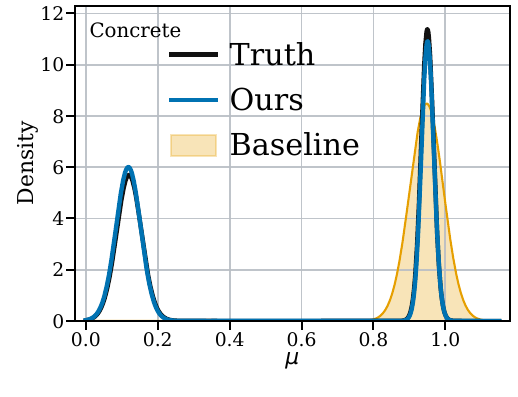}
\label{fig:concrete_overlay}
\end{subfigure}
\hfill
\begin{subfigure}[t]{0.49\linewidth}
\centering
\includegraphics[width=\linewidth]{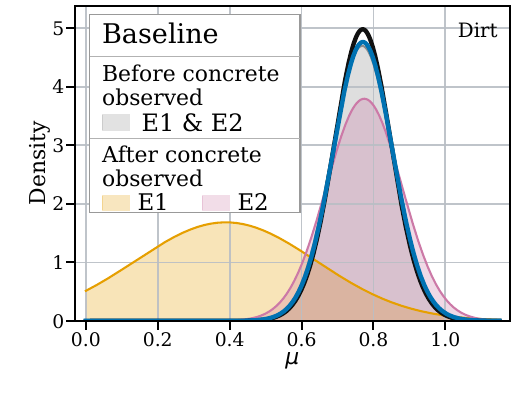}
\label{fig:dirt_overlay}
\end{subfigure}

\vspace*{-5mm}

\begin{subfigure}[t]{0.49\linewidth}
\centering
\includegraphics[width=\linewidth]{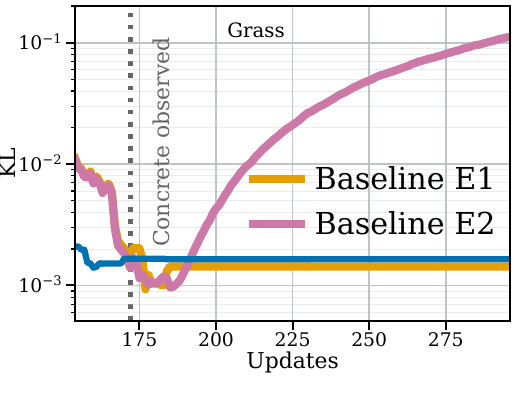}
\label{fig:kl_grass}
\end{subfigure}
\hfill
\begin{subfigure}[t]{0.49\linewidth}
\centering
\includegraphics[width=\linewidth]{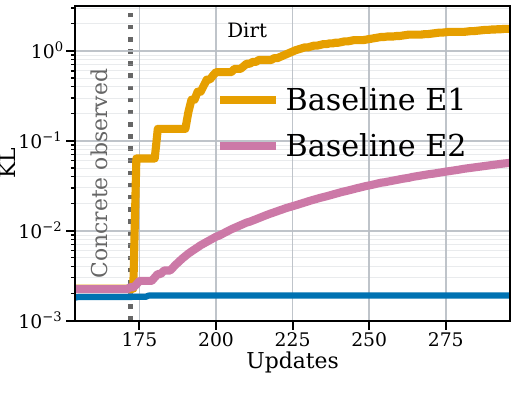}
\label{fig:kl_dirt}
\end{subfigure}
\vspace{-6mm}
\caption{Multimodality and class contamination}
\label{fig:exp1}
\vspace{-3mm}
\end{wrapfigure}

The vehicle follows a fixed trajectory that first traverses grass, then dirt, and then concrete. Both scenarios use the same trajectory and the same interaction-derived friction estimates. For each displayed material, we evaluate one representative prediction against the corresponding ground-truth target. Our method queries the empirical mean embedding of that material’s observations, while the semantic baseline uses its class label. The semantic baseline aggregates visual evidence within regions whose size and boundaries must first be defined or estimated. Determining these regions can be nontrivial, and their boundaries affect the evidence being aggregated. We provide ground-truth region boundaries in these experiments. Our estimator instead retrieves and weights individual interaction beliefs directly from a query embedding, avoiding this additional region-definition step.

\noindent\textbf{Multimodality failure.}
When concrete is a known class but has two friction regimes, the semantic baseline cannot represent the bimodal concrete target with a single class-level belief. It captures the high-friction mode but misses the low-friction mode, as shown in Figure~\ref{fig:exp1} top left. In contrast, our estimator retains the high- and low-friction concrete measurements as separate mixture components and recovers both modes. The concrete measurements can also affect unrelated semantic classes. The baseline shifts the dirt belief to explain low-friction concrete measurements, causing a previously correct dirt estimate to move away from its target, as illustrated in Figure~\ref{fig:exp1} top right and Figure~\ref{fig:exp1} bottom right. Appendix~\ref{app:multimodal_risk} illustrates how collapsing these modes can underestimate collision risk.

\noindent\textbf{Out-of-vocabulary contamination.}
When concrete is absent from the semantic vocabulary, its measurements must be routed through the grass and dirt labels. Figures~\ref{fig:exp1} bottom left and right show the KL divergence over the update sequence, with the dotted line marking the first concrete contact. The grass and dirt predictions match their targets before concrete is observed. After concrete measurements begin, the baseline KL divergences increase for the known unimodal classes, while our estimator remains stable. 

\par
\ifnum\value{WF@wrappedlines}>1
  \vspace{\dimexpr\value{WF@wrappedlines}\baselineskip-\baselineskip\relax}
\fi
\WFclear
\vspace*{-2mm}
\subsection{Kernel Flexibility and Runtime Scaling}

\paragraph{Composing visual and spatial similarity.}
\begin{wrapfigure}{l}{3.2in}
\centering
\vspace*{-4mm}
\includegraphics[width=\linewidth]{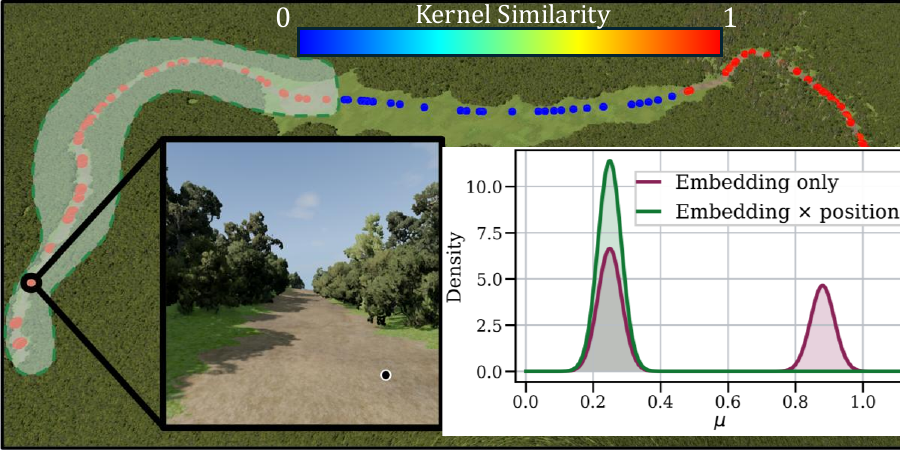}
\vspace*{-5mm}
\caption{Composite kernels combine visual similarity and spatial proximity for localized history retrieval.}
\label{fig:flexible_kernel}
\vspace*{-3mm}
\end{wrapfigure}
We illustrate that the retrieval kernel need not depend only on visual embeddings. In this experiment, the vehicle collects estimates along a route containing two visually similar dirt segments separated by a grassy valley. The two dirt segments share a similar appearance but correspond to different physical response distributions. For a query on the first dirt segment, an embedding-only kernel retrieves interaction history from both dirt regions, producing a bimodal posterior.
To incorporate spatial locality, we compose the embedding kernel with a position kernel, $K(e_q,e_i,p_q,p_i) = K_e(e_q,e_i) \exp(-|p_q-p_i|_2^2 / (2\ell_p^2))$ where $p_q$ and $p_i$ are the query and history positions, and $\ell_p$ is the spatial bandwidth. As shown in Fig.~\ref{fig:flexible_kernel}, this kernel suppresses the distant dirt while preserving nearby visually-similar history, yielding a unimodal distribution.\vspace*{-2mm}

\par
\ifnum\value{WF@wrappedlines}>1
  \vspace{\dimexpr\value{WF@wrappedlines}\baselineskip-\baselineskip\relax}
\fi
\WFclear
\FloatBarrier
\medskip

\Needspace{12\baselineskip}
\begin{wraptable}{r}{0.53\linewidth}
\vspace{-2mm}
\centering
\footnotesize
\caption{Sampling runtime for $S$ 21-dimensional posterior samples with history size $N$. We report CPU/GPU times (ms).}

\label{tab:history_query_runtime}
\setlength{\tabcolsep}{4.2pt}
\begin{tabular}{c c c c c}
\toprule
$N$ & $S=10$ & $S=100$ & $S=1000$ \\
\midrule
$10^4$ & 0.51 / 0.68 & 0.54 / 0.68 & 0.81 / 0.74 \\
$10^5$ & 3.62 / 2.06 & 4.11 / 2.07 & 4.01 / 2.07 \\
$10^6$ & 51.61 / 15.39 & 51.45 / 15.36 & 49.84 / 15.22 \\
\bottomrule
\end{tabular}
\vspace{-2mm}
\end{wraptable}

\noindent
\textbf{Runtime scaling with interaction history.}
We benchmark how history-query latency scales with the number of stored interactions.
Query time grows with database size but remains small in practice. With $10^6$
entries, the GPU implementation returns $1000$ posterior samples in $15.22\,\mathrm{ms}$. The weak dependence on sample count reflects that the dominant cost is computing similarities to the database, after which drawing samples from the mixture is inexpensive.

\medskip




\vspace*{-2mm}
\subsection{Closed-Loop Racing with Vision-Based Online Physical-Parameter Estimation}
\label{subsec:live_trajopt_mpc}
\vspace*{-2mm}

We evaluate whether the proposed posterior can improve closed-loop control in a setting where surface-dependent parameters directly affect the vehicle's feasible speed and handling limits. We integrate the estimator with a long-horizon trajectory optimizer and a short-horizon MPC tracker in BeamNG. Both controllers use a nonlinear bicycle model~\cite{Paden2016ASO,rajamani2011vehicle, Falcone2007PredictiveAS,Kong2015KinematicAD,Polack2017TheKB} in Frenet coordinates with respect to a reference path~\cite{ Lew2024RiskAverseMP, Subosits2021ImpactsOM}. The state and input are
$x=[s,t,r,V,\beta,\omega_f,\omega_r,e,\Delta\psi,\delta,T_{\mathrm{eng}}],
u=[\dot\delta_{\mathrm{cmd}},\dot T_{\mathrm{eng,cmd}}] ,
$
where $s$ is the distance along the reference path, $t$ is elapsed time, $r$ is yaw rate, $V$ is speed, $\beta$ is sideslip, $\omega_f$, $\omega_r$ are wheel-speed states, $e$ and $\Delta\psi$ are lateral and heading errors, and $\delta,T_{\mathrm{eng}}$ are steering angle and engine torque. The dynamics are written compactly as
$\dot x = f_{\mathrm{dyn}}(x,u,\mathcal{F}_{\theta}(x,u))$ with $\mathcal{F}_{\theta}=[F^\theta_{xf},F^\theta_{yf},F^\theta_{xr},F^\theta_{yr}]$
where $\mathcal{F}_{\theta}$ is a learned $\mathrm{ExpTanh}$ tire-force model~\cite{djeumou2023autonomous} trained offline from BeamNG data collected across multiple surface conditions. In this experiment, the $21$-latent physical parameters estimated by our method are the tire-model tune vector, i.e., $z\equiv\theta$, which modulates slip recentering, slip weighting, damping, tanh sharpness, force gains, and surface-dependent terms. Full model, parameter, and solver details are provided in Appendix~\ref{app:controls}.


\noindent\textbf{Surface-aware trajectory generation.} At each replanning cycle, the latest reference trajectory defines the preview locations at which the estimator is queried. Points ahead of the vehicle along this reference are associated with visual observations, and the estimator returns posterior beliefs over the corresponding surface-dependent tire parameters. We use the posterior mean at each preview point to form the stagewise tune matrix $P_\theta=[\theta_0,\ldots,\theta_{H-1}]$, which parameterizes the tire-force model along the long-horizon trajectory-optimization problem:
\begin{align}
\min_{x_k,u_k} \quad
& t_H + \sum\nolimits_{k=0}^{H-1} u_k^\top R u_k \\
\text{s.t.} \quad
& x_{k+1} - x_k
= F_s(x_{k+1},u_k,\kappa_k,\theta_k), \\
& g(x_k,u_k,\theta_k) \le 0,
\quad \forall k.
\label{eq:vision_informed_trajopt}
\end{align}
Here $F_s$ denotes the spatially normalized dynamics obtained from $dx/ds=\dot{x}/\dot{s}$, $\kappa_k$ is the path curvature, and $\theta_k$ is the surface-dependent tire-model parameter inferred for the $k$th preview point. The optimizer periodically re-solves this problem and publishes an updated surface-aware reference. As the vehicle tracks the reference, newly traversed regions provide interaction measurements that are converted into local parameter beliefs and added to $\mathcal{D}_N$, so the next optimization is informed by both pre-contact vision and accumulated interaction evidence. A higher-rate MPC tracks the latest optimized reference using the same vehicle model, but a short-horizon quadratic tracking objective. 

\begin{figure*}[!t] 
\centering 
\includegraphics[width=0.95\textwidth]{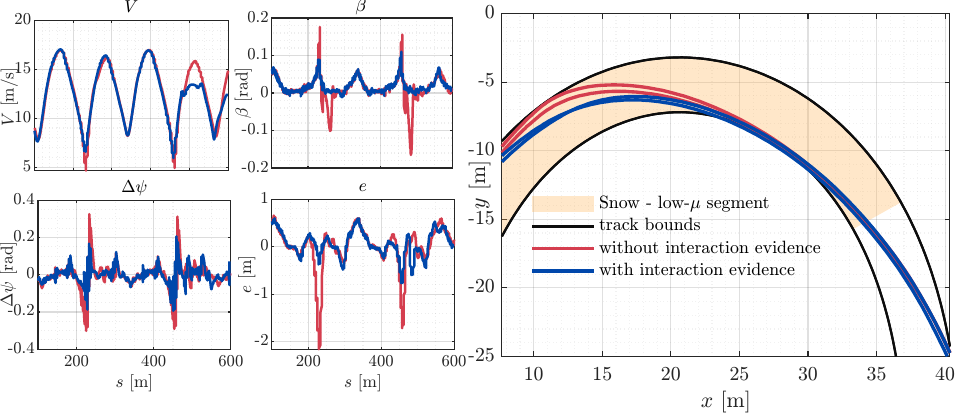}
\vspace*{-2mm}
\caption{Trajectory optimization solution and closed-loop tracking comparison over the snow--low-$\mu$ segment. The left panels show the evolution of selected vehicle states along the path, while the right panel compares the planned trajectory with and without interaction-based estimation.}
\vspace*{-3mm}
\label{fig:traj_opt_sol}
\end{figure*}

\textbf{Closed-loop adaptation on mixed-friction surfaces.}
We test this stack on an oval track containing high-friction asphalt and an unknown low-friction icy segment. A generative prior-only controller, which does not update tire parameters from interaction, plans aggressively into the low-friction region, typically reaches the lateral-error bounds, and exhibits peak entry errors near $2\,\mathrm{m}$ with sideslip increasing to about $0.18\,\mathrm{rad}$. In contrast, the proposed estimator updates the posterior as interaction evidence accumulates, allowing the optimizer to generate progressively more surface-aware references. As shown in Fig.~\ref{fig:traj_opt_sol}, online adaptation reduces the low-friction entry error to about $0.5\,\mathrm{m}$ and keeps sideslip below $0.1\,\mathrm{rad}$, significantly improving closed-loop racing performance relative to baseline.

Across nine complete traversals of the same $s=200$--$300\,\mathrm{m}$ low-friction segment, the proposed estimator reduces traversal time from $9.304\!\pm\!0.316$ to $8.135\!\pm\!0.007\,\mathrm{s}$ ($12.6\%$) and increases mean speed from $10.51$ to $11.78\,\mathrm{m/s}$. The prior-only baseline overshoots at ice entry and slows to $3.0$--$4.8\,\mathrm{m/s}$ during MPC correction, whereas the interaction-based controller remains above $7.84\,\mathrm{m/s}$ with $16\times$ lower across-traversal speed-profile variance. Lateral-error, heading-error, and sideslip root-mean-square errors decrease from $1.637$ to $0.121\,\mathrm{m}$, $0.173$ to $0.017\,\mathrm{rad}$, and $0.089$ to $0.027\,\mathrm{rad}$, respectively. These results show that a more conservative planned entry speed can preserve momentum and reduce traversal time. Additional plots and implementation details are provided in Appendix~\ref{app:controls}.

\begin{figure*}[!t]
\centering
\includegraphics[width=\textwidth]{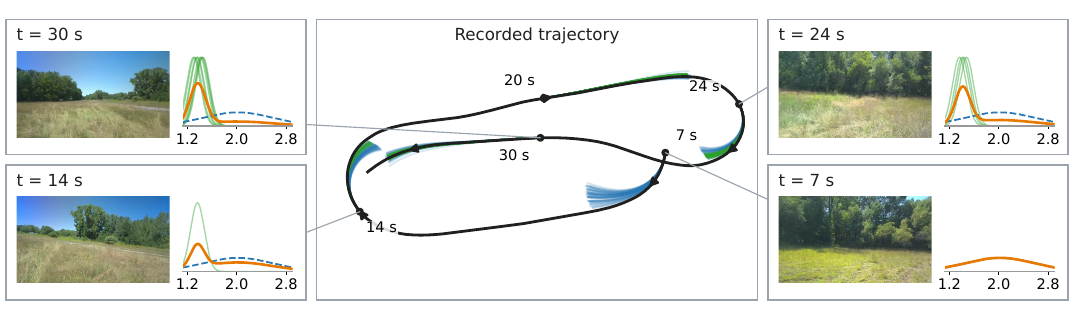}
\vspace{-2mm}
\caption{Recorded real-vehicle trajectory (black) and rollout clouds
from the prior (blue) and retrieved history (green).
Insets show look-ahead images and front lateral-stiffness beliefs:
prior (blue dashed), history Gaussians (green),
and fused belief (orange).}
\label{fig:hardware_rollouts}
\vspace{-2mm}
\end{figure*}

\subsection{Real-vehicle parameter prediction}
\label{sec:hardware}

We evaluate vehicle-motion prediction on recorded grass trajectories
from the autonomous off-road vehicle shown in
Appendix~\ref{app:hardware}. Local Gaussian parameter beliefs
are inferred from measured states and controls through trajectory-based
nonlinear log-likelihood inversion, with covariances obtained from a
temperature-scaled Gauss--Newton inverse-Hessian approximation.
These beliefs are stored with the corresponding terrain embeddings.
Embedding--measurement pairs from a separate concrete run are included
as distractors, testing retrieval of relevant grass evidence.

Retrieval uses a cosine-similarity cutoff of $0.90$, kernel scale $10$,
and support parameter $\kappa=2$. We set $\alpha=0$.
At selected locations, look-ahead images supply query embeddings.
We sample physical parameters from the prior and retrieved history
and simulate $2.5\,\mathrm{s}$ trajectories from the same measured
state under the recorded controls. This evaluates prediction against
the recorded trajectory rather than closed-loop adaptation.

Figure~\ref{fig:hardware_rollouts} compares the resulting rollouts.
As interaction evidence accumulates across $18$ measurements, mean
per-particle average displacement error over the prediction horizon
decreases from $0.737$ to $0.628\,\mathrm{m}$ ($14.8\%$), while final
displacement error decreases from $2.096$ to $1.750\,\mathrm{m}$
($16.5\%$).

\vspace*{-2mm}
\section{Conclusion}
\label{sec:conclusion}
\vspace*{-2mm}

We propose an interaction-grounded estimator for online physical-parameter prediction from visual observations. By combining continuous visual embeddings with Gaussian beliefs inferred from prior interactions, the method produces an explicit multimodal posterior without semantic class indexing. Through high-fidelity simulation, we show that the estimator avoids class-level averaging and out-of-vocabulary contamination, scales to large interaction histories at real-time control rates, and enables closed-loop adaptation in autonomous racing across low- and high-friction surfaces.

\textbf{Limitations and future work.}
Closed-loop adaptation has so far been evaluated in simulation; the real-vehicle experiment evaluates parameter prediction under recorded controls. In addition, although the estimator represents a multimodal posterior over physical parameters, the current controller uses posterior summaries rather than explicitly optimizing over the full distribution. Risk-sensitive or chance-constrained control could further improve robustness and guide active data collection in poorly supported regions. Finally, our formulation assumes access to local Gaussian beliefs from interaction; improving the robustness of trajectory-based inverse modeling across broader operating conditions remains an important direction for future work.


\clearpage
\acknowledgments{This work was supported by Toyota Research Institute.}


\bibliography{example}

\clearpage
\appendix
\input{appendix}

\end{document}

%% file: appendix.tex
\noindent\textbf{Supplementary video.}
A supplementary video overview is available at:
\url{https://tinyurl.com/vision-param-estimation}.

\vspace{1em}

\section{Visual Embedding Refinement Details}
\label{app:embedding_refinement}

This appendix gives the implementation details for the retrieval-space refinement described in Sec.~3.

\subsection{Feature extraction}

We use a frozen DINOv3 ViT-B/16 encoder as the base visual encoder. The encoder receives ImageNet-normalized images and produces a $32 \times 32$ grid of 768-dimensional patch-token embeddings. Prefix tokens are discarded, and only patch tokens are used as local region descriptors.

Before refinement, each patch token is debiased using the positional-subspace projection of INSID3~\cite{insid3} and then normalized. We estimate a 256-dimensional positional basis from one Gaussian-noise image at the same image resolution as the rendered observations. Let $U$ denote this basis. For a raw patch token $x$, the base feature is
\begin{equation}
\bar e = \mathrm{unit}(x - UU^\top x),
\end{equation}
where $\mathrm{unit}(v)=v/|v|$. The basis is saved after estimation and reused for data preparation, training, visualization, and deployment.

\subsection{Prototype discovery}

The refinement model is trained on an image corpus collected in BeamNG from a flat square map whose tiles are assigned random terrain types. This map is separate from the deployment scenes used in the main experiments, which contain additional visual structure such as trees and hills. The corpus contains 1,500 images, giving 1,536,000 patch embeddings.

We implement the density-based clustering step from Sec.~3 using HDBSCAN. We first sample 100,000 patch embeddings uniformly from the full database and reduce them to 64 dimensions with PCA. HDBSCAN is run in the PCA space with Euclidean distance, minimum cluster size 500, and minimum samples 10. The run used throughout the experiments found $K=7$ clusters, with 35,693 of the sampled patches marked as noise.

Prototype directions are computed in the original 768-dimensional feature space. If $a_i$ is the HDBSCAN assignment for sampled feature $\bar e_i$, the prototype for cluster $k$ is
\begin{equation}
c_k =
\mathrm{unit}
\left(
\sum_{i:a_i=k} \bar e_i
\right).
\end{equation}
After computing these prototypes, every patch in the full feature database is assigned to its nearest prototype by cosine similarity. Patches whose maximum cosine similarity to any prototype is below 0.5 are assigned as noise. This assigns 1,532,716 of 1,536,000 patches and leaves 3,284 patches as noise.

\subsection{Residual model details}

The residual model is an MLP $r_\eta$ from 768-dimensional base features to 768-dimensional residuals. It has three hidden layers, hidden width 1024, and ReLU activations. The model is trained with Adam for 10,000 steps using batches of 256 assigned patches and 256 noise patches. We use learning rate $10^{-3}$, $\lambda_{\mathrm{nonnoise}}=0.1$, and $\lambda_{\mathrm{noise}}=0.0005$. Training was run on an NVIDIA RTX 4070 laptop GPU.

\subsection{Qualitative refinement visualization}

\begin{figure}[t]
\centering
\includegraphics[width=\linewidth]{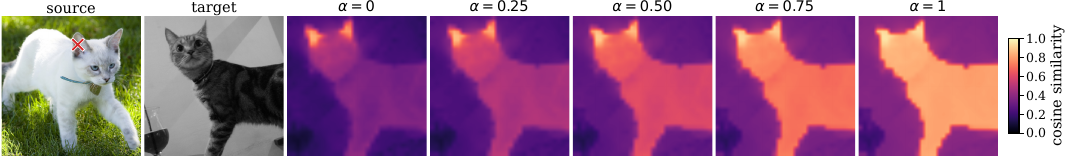}
\caption{
Qualitative visualization of the residual refinement on a cat-image dataset.
}
\label{fig:cat_refinement}
\end{figure}

We also apply the same prototype-discovery and residual-refinement procedure to a cat-image dataset to visualize the learned retrieval geometry. The cat refinement model uses the same architecture, clustering procedure, and training hyperparameters as the BeamNG refinement model, except that the non-noise residual penalty is set to $\lambda_{\mathrm{nonnoise}}=1$. Fig.~\ref{fig:cat_refinement} shows the cosine similarity induced by a selected query patch as the refinement strength $\alpha$ is increased. As $\alpha$ increases, the similarity induced by the ear query becomes stronger around the corresponding ear regions.

%
%

\section{Simulation and Experimental Setup}
\label{app:sim_details}

This appendix details the simulation environment and the runtime pipeline shared across all simulation experiments, followed by the hyperparameters used in each. The controller and generative-prior details specific to the closed-loop racing experiment are deferred to Appendix~\ref{app:controls}.

\subsection{Environment}
\label{app:sim_details:env}

\begin{figure}[!hbt]
    \centering
    \includegraphics[width=0.95\linewidth]{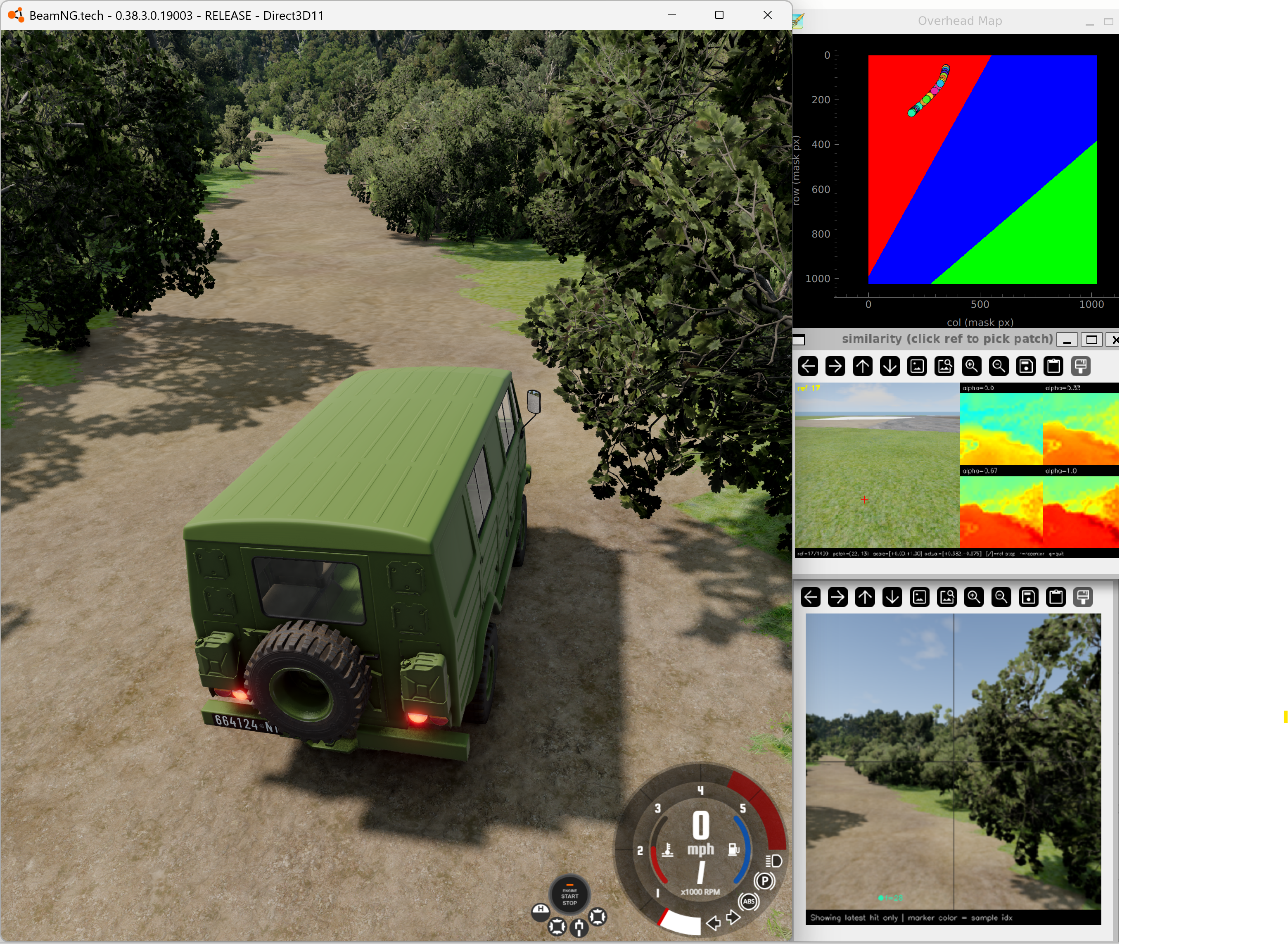}
    \caption{The runtime visualizer during a live drive. \emph{Left:} the BeamNG.tech simulator showing the vehicle traversing a dirt road. \emph{Top right:} the overhead map of the known per-pixel terrain regions, overlaid with the vehicle's trajectory; each marker is a contact event whose world position indexes the map to assign a surface and draw an interaction belief. \emph{Middle right:} a reference image (left) with a selected query patch (red cross), and the cosine-similarity maps it induces over the current camera frame at refinement strengths $\alpha \in \{0, 0.33, 0.67, 1.0\}$; \emph{Bottom right:} Visualization of the frame in the rolling buffer which the most recent contact event projected to.}
    \label{fig:setup_visualizer}
\end{figure}

All simulation experiments run in the BeamNG.tech driving simulator~\cite{beamngdrive}, which provides high-fidelity vehicle dynamics and rendered camera imagery. A forward-facing pinhole camera rigidly mounted to the vehicle produces $512\times512$ RGB frames; its calibrated intrinsics $K$, mounting offset, and per-frame pose $T_{wc}^{(f)}\in SE(3)$ are recorded with each frame so that world points can later be projected into image space. The simulator reports the vehicle's world-frame position at every step, and the terrain layout is known as a per-pixel map, so the surface a contact lands on is recovered by indexing the map at the vehicle's position; this terrain map is used only to assign the physical surface at a contact, never as an input to the estimator. Fig.~\ref{fig:setup_visualizer} shows the live system during a run.

\subsection{Runtime pipeline}
\label{app:sim_details:pipeline}

We describe the pipeline in the order events occur during a run: frames are encoded as they arrive, buffered with their poses, and later associated with contact events to populate the interaction history $\mathcal{D}_N$.

\paragraph{Encoding.}
Each incoming camera frame is passed through a single perception pipeline that is reused identically at training, deployment, and offline replay. The pipeline (i) encodes the frame with a frozen DINOv3 ViT-B/16 backbone~\cite{Simeoni2025DINOv3}, producing a $32\times32$ grid of $768$-dimensional patch embeddings (ii) removes a fixed low-dimensional positional component from each patch embedding (iii) applies the learned residual refinement $r_\eta$ of Sec.~\ref{sec:refinement} and (iv) renormalizes each patch embedding to the unit sphere. The first two stages are deterministic and shared with the embedding refinement training described in Appendix~\ref{app:embedding_refinement}.

\paragraph{Frame buffering.}
Encoded frames are placed in a rolling buffer that retains the most recent $256$ frames together with their camera poses. Buffering is necessary because contact evidence is not available at the instant a region is first seen: the vehicle observes a region from a distance, then contacts it some frames later. The buffer therefore preserves a short window of recent viewpoints so that, when a contact occurs, the system can recover a frame in which the contacted location was actually visible.

\paragraph{Contact events and projection.}
As the vehicle drives, contact events are emitted along its trajectory at a controlled rate, each corresponding to a world-frame location $p\in\mathbb{R}^3$ the vehicle has just traversed. To associate $p$ with a visual embedding, we project it into every buffered frame $f$. Writing the point in the camera frame as $p^{(f)}_{c}=R_{\mathrm{cam}}\,(T_{wc}^{(f)})^{-1}\,\tilde p$ (with $\tilde p$ the homogeneous lift of $p$ and $R_{\mathrm{cam}}$ the fixed rotation between the simulator and pinhole camera conventions), the pixel coordinate is $u^{(f)}=\pi\!\left(K\,p^{(f)}_{c}\right)$, where $\pi$ denotes the perspective division. A frame is valid only if $p$ lies in front of the camera and within the image, i.e. $[p^{(f)}_{c}]_z>0$ and $u^{(f)}\in[0,W)\times[0,H)$. Among valid frames we select the one minimizing $\lVert u^{(f)}-u_{\mathrm{center}}\rVert$.

\paragraph{Pairing and the interaction history.}
The same contact location indexes the terrain map to recover the surface at $p$, and that surface's model returns a Gaussian belief $\phi_i=(\mu_i,\Sigma_i)$ over the local physical parameter (Sec.~\ref{app:sim_details:nig}). The pair $(e_i,\phi_i)$ is appended to the interaction history $\mathcal{D}_N$, which stores embeddings on the GPU for the cosine-similarity matmul and the belief parameters alongside them. At query time, the support-aware mixture of Sec.~\ref{sec:method} is evaluated against the current contents of $\mathcal{D}_N$. 

\subsection{Inverse-model abstraction}
\label{app:sim_details:nig}

To remain agnostic to any particular inverse-modeling procedure, the simulator supplies the local Gaussian belief directly rather than fitting a raw trajectory. Each terrain region is assigned a Normal--Inverse--Gamma (NIG) model over the physical parameter; at a contact, the simulator draws a belief $\phi=(\mu,\Sigma)$ from that region's NIG, which plays the role of the output a calibrated inverse model would return from a short interaction. Per dimension, with hyperparameters $(\mu_0,\lambda,\alpha,\beta)$,
\begin{equation*}
    \Sigma \sim \mathrm{InverseGamma}(\alpha,\beta),
    \qquad
    \mu \mid \Sigma \sim \mathcal{N}\!\left(\mu_0,\,\Sigma/\lambda\right),
\end{equation*}
so $\mu_0$ sets the expected parameter value, $\lambda$ controls how tightly the mean concentrates, and $(\alpha,\beta)$ govern the spread of the belief. When a region is physically heterogeneous, such as concrete interspersed with ice, it is assigned multiple NIG components and the simulator draws a component uniformly at each contact before sampling.

\subsection{Hyperparameters}
\label{app:sim_details:hyperparams}

\paragraph{Surface friction models.}
Table~\ref{tab:setup_nig_params} lists the NIG hyperparameters for the scalar-friction experiments. The $21$-dimensional tire-parameter models used in the racing experiment are described in Appendix~\ref{app:controls}.

\begin{table}[!hbt]
\centering
\footnotesize
\setlength{\tabcolsep}{6pt}
\begin{tabular}{l l l c c c c}
\toprule
Experiment & Region & Physical interpretation & $\mu_0$ & $\sigma$ & $\lambda$ & $\alpha$ \\
\midrule
\multirow{4}{*}{E1 (bimodal concrete)}
  & grass         & moderate friction           & 0.86 & 0.09  & 30 & 25 \\
  & dirt          & moderate friction           & 0.77 & 0.08  & 30 & 25 \\
  & concrete (hi) & dry concrete, high $\mu$     & 0.95 & 0.018 & 30 & 25 \\
  & concrete (lo) & ice patch, very low $\mu$    & 0.12 & 0.035 & 30 & 25 \\
\midrule
\multirow{3}{*}{E2 (OOV concrete)}
  & grass         & moderate friction           & 0.86 & 0.09  & 30 & 25 \\
  & dirt          & moderate friction           & 0.77 & 0.08  & 30 & 25 \\
  & concrete      & dry concrete (out-of-vocab) & 0.95 & 0.018 & 30 & 25 \\
\midrule
\multirow{3}{*}{Kernel-similarity exp.}
  & dirt A        & query surface               & 0.25 & 0.035 & 60 & 35 \\
  & dirt B        & visual twin, different $\mu$ & 0.88 & 0.035 & 60 & 35 \\
  & grass         & neutral (off-query)         & 0.55 & 0.050 & 60 & 35 \\
\bottomrule
\end{tabular}
\caption{Per-region NIG hyperparameters for the friction experiments}
\label{tab:setup_nig_params}
\end{table}

\paragraph{Retrieval kernel.}
All experiments use the cosine-similarity exponential kernel $K_h(e_q,e_i)=\exp\!\bigl(\gamma(e_q^\top e_i-1)\bigr)\,\mathbf{1}[e_q^\top e_i>c]$, where $\gamma$ sets retrieval sharpness, the support cutoff $\kappa$ sets how much accumulated similarity is required before the history is trusted over the prior, and $c$ is an optional hard similarity floor. Experiment~1 uses $\gamma=400$ and $\kappa=5\times10^{-3}$. Experiment~2 uses an embedding sharpness $\gamma=40$ and, for the composite kernel, a spatial bandwidth $\ell_p=170$, with $\kappa=5\times10^{-3}$. The racing experiment uses $\gamma=100$, $\kappa=10^{-2}$, and a hard floor $c=0.9$.

\section{Vision-Informed Control Stack Details}
\label{app:controls}

This appendix provides implementation details for the closed-loop vision-informed racing experiment in Sec.~\ref{subsec:live_trajopt_mpc}. The control stack combines a real-time long-horizon trajectory optimizer, a short-horizon MPC tracker, and an online vision estimator that supplies stagewise tire-model parameters. The optimizer and MPC use the same Frenet-frame state and input,
$
x=[s,t,r,V,\beta,\omega_f,\omega_r,e,\Delta\psi,\delta,T_{\mathrm{eng}}]^\top
$
and
$
u=[\dot{\delta}_{\mathrm{cmd}},\dot{T}_{\mathrm{eng,cmd}}]^\top
$.

\subsection{Vehicle Dynamics}
\label{app:vehicle_dynamics}

The optimizer and MPC use a dynamic bicycle model augmented with Frenet-frame path states ~\cite{Paden2016ASO,rajamani2011vehicle, Falcone2007PredictiveAS,Kong2015KinematicAD,Polack2017TheKB,Lew2024RiskAverseMP, Subosits2021ImpactsOM}. With
$x=[s,t,r,V,\beta,\omega_f,\omega_r,e,\Delta\psi,\delta,T_{\mathrm{eng}}]^\top$ and
$u=[\dot{\delta}_{\mathrm{cmd}},\dot T_{\mathrm{eng,cmd}}]^\top$, the continuous-time dynamics are
\begin{equation}
\dot{x}=
\begin{bmatrix}
\dot{s}\\ \dot{t}\\ \dot{r}\\ \dot{V}\\ \dot{\beta}\\
\dot{\omega}_f\\ \dot{\omega}_r\\ \dot{e}\\
\dot{\Delta\psi}\\ \dot{\delta}\\ \dot{T}_{\mathrm{eng}}
\end{bmatrix}
=
\begin{bmatrix}
\dfrac{V\cos\Delta\psi}{1-e\kappa(s)}\\[0.35em]
1\\[0.35em]
\dfrac{aF^\theta_{yf}\cos\delta+aF^\theta_{xf}\sin\delta-bF^\theta_{yr}}{I_z}\\[0.6em]
\dfrac{-F^\theta_{yf}\sin(\delta-\beta)+F^\theta_{xf}\cos(\delta-\beta)
+F^\theta_{yr}\sin\beta+F^\theta_{xr}\cos\beta}{m}\\[0.6em]
\dfrac{F^\theta_{yf}\cos(\delta-\beta)+F^\theta_{xf}\sin(\delta-\beta)
+F^\theta_{yr}\cos\beta-F^\theta_{xr}\sin\beta}{mV}-r+\Delta_{\beta,\theta}\\[0.6em]
c_f\!\left(-F^\theta_{xf}-c_{bf}m b_r\right)\\[0.35em]
c_TT_{\mathrm{eng}}+c_r\!\left(-F^\theta_{xr}-c_{br}m b_r\right)\\[0.35em]
V\sin\Delta\psi\\[0.35em]
\dot{\beta}+r-\kappa(s)\dot{s}\\[0.35em]
\dot{\delta}_{\mathrm{cmd}}\\[0.35em]
\dot T_{\mathrm{eng,cmd}}
\end{bmatrix}.
\label{eq:full_vehicle_dynamics_app}
\end{equation}
Here $m,I_z,a,b$ are the vehicle mass, yaw inertia, and axle distances, $\kappa(s)$ is the path curvature, and
$\mathcal{F}_{\theta}=[F^\theta_{xf},F^\theta_{yf},F^\theta_{xr},F^\theta_{yr}]^\top$ is the learned tire-force model evaluated with the surface-dependent tune vector $\theta$. The brake input $b_r$ is not optimized in the trajectory-optimization runs and is set to zero. This model relied on engine braking. The auxiliary velocity and slip quantities used by the tire model are
\begin{equation}
\begin{gathered}
v_x=V\cos\beta,\qquad v_y=V\sin\beta,\\
v_{xf}=v_x\cos\delta+(v_y+a r)\sin\delta,\qquad
v_{yf}=-v_x\sin\delta+(v_y+a r)\cos\delta,\\
v_{xr}=v_x,\qquad v_{yr}=v_y-b r,\\
\kappa_{xf}=\frac{\omega_f-v_{xf}}{v_{xf}},\qquad
\alpha_f=\mathrm{atan2}(v_y+a r,v_x)-\delta,\\
\kappa_{xr}=\frac{\omega_r-v_{xr}}{v_{xr}},\qquad
\alpha_r=\mathrm{atan2}(v_{yr},v_{xr}).
\end{gathered}
\label{eq:auxiliary_slip_terms_app}
\end{equation}
The spatial model used by the trajectory optimizer is obtained through the chain rule $\frac{dx}{ds}=\frac{\dot{x}}{\dot{s}}$.
In addition to modifying the tire-force map, four tune parameters directly augment the sideslip dynamics through
\begin{equation}
\Delta_{\beta,\theta}
=
c_{\beta,\dot\alpha f}\dot{\alpha}_f+
c_{\beta,\dot\alpha r}\dot{\alpha}_r+
c_{\beta,\dot\kappa f}\dot{\kappa}_{xf}+
c_{\beta,\dot\kappa r}\dot{\kappa}_{xr}.
\label{eq:beta_residual_app}
\end{equation}
These terms act as a low-dimensional residual model for unmodeled sideslip dynamics, helping account for high-CG and rapid load-transfer effects observed during aggressive all-terrain vehicle maneuvers in the high-fidelity BeamNG simulation.

\subsection{Learned ExpTanh Tire Model and Tune Parameters}
\label{app:tire_tune_params}

The tire forces are represented by a learned ExpTanh model~\cite{djeumou2023autonomous}. For the front and rear tires, define
\begin{equation}
\begin{aligned}
S_f &=
\sqrt{
\left[k_{\kappa f}\left(\kappa_{xf}-(s_{fx}+d s_{fx})\right)\right]^2
+\mu_f c_f\left[\alpha_f-(s_{fy}+d s_{fy})\right]^2+\epsilon},\\
S_r &=
\sqrt{
\left[k_{\kappa r}\left(\kappa_{xr}-(s_{rx}+d s_{rx})\right)\right]^2
+\mu_r c_r\left[\alpha_r-(s_{ry}+d s_{ry})\right]^2+\epsilon}.
\end{aligned}
\label{eq:tuned_total_slip_app}
\end{equation}
The damping factors are $d_f=\exp(-\rho_f\lambda_f S_f)$ and $d_r=\exp(-\rho_r\lambda_r S_r)$, and the learned features are passed through
\begin{equation}
z_f=\tanh\!\left(\frac{g_f(\cdot)}{\tau_f}\right)\odot d_f,
\qquad
z_r=\tanh\!\left(\frac{g_r(\cdot)}{\tau_r}\right)\odot d_r.
\label{eq:exptanh_app}
\end{equation}
The final tire forces are
\begin{equation}
\begin{bmatrix}F^\theta_{xf}\\F^\theta_{yf}\end{bmatrix}
=
m
\begin{bmatrix}
\gamma_{xf} & 0\\
0 & \gamma_{yf}
\end{bmatrix}
h_f(z_f),
\qquad
\begin{bmatrix}F^\theta_{xr}\\F^\theta_{yr}\end{bmatrix}
=
m
\begin{bmatrix}
\gamma_{xr} & 0\\
0 & \gamma_{yr}
\end{bmatrix}
h_r(z_r).
\label{eq:tire_force_map_app}
\end{equation}
Here $g_f,g_r$ are learned feature maps and $h_f,h_r$ are learned force-output maps; all four are trained offline and converted to CasADi functions for real-time optimization. The canonical tune vector is $\theta\in\mathbb{R}^{21}$. Most entries modulate the tire-force model, while four entries directly augment the sideslip dynamics as in~\eqref{eq:beta_residual_app}. Log-parameters are exponentiated before use, so zero corresponds to a neutral multiplicative factor of one. The optimizer receives a stagewise tune matrix $P_\theta=[\theta_0,\theta_1,\ldots,\theta_{N_{\mathrm{TO}}-1}],
~
\theta_k\in\mathbb{R}^{21}$, where each column corresponds to the posterior-mean tire/dynamics tune vector inferred for one preview point along the currently planned trajectory.

\begin{table*}[!t]
\centering
\caption{Canonical tune vector used by the vision-informed trajectory optimizer.}
\label{tab:tune_params_app}
\footnotesize
\setlength{\tabcolsep}{5pt}
\renewcommand{\arraystretch}{1.08}
\begin{tabular}{lll}
\toprule
\textbf{Group} & \textbf{Parameters} & \textbf{Role} \\
\midrule
Sideslip residual dynamics &
\shortstack[l]{$c_{\beta,\dot{\alpha}f},\,c_{\beta,\dot{\alpha}r}$\\
$c_{\beta,\dot{\kappa}f},\,c_{\beta,\dot{\kappa}r}$} &
Residual correction to $\dot{\beta}$. \\

Slip recentering &
$d s_{fx},\,d s_{fy},\,d s_{rx},\,d s_{ry}$ &
Shift slip centers. \\

Force gains &
\shortstack[l]{$\log\gamma_{xf},\,\log\gamma_{yf}$\\
$\log\gamma_{xr},\,\log\gamma_{yr}$} &
Scale tire forces. \\

Slip scaling/weighting &
\shortstack[l]{$\log k_{\kappa f},\,\log k_{\kappa r}$\\
$\log\mu_f,\,\log\mu_r$} &
Tune slip sensitivity. \\

Damping strength &
$\log\rho_f,\,\log\rho_r$ &
Scale slip damping. \\

Tanh sharpness &
$\log\tau_f,\,\log\tau_r$ &
Set ExpTanh sharpness. \\

Compatibility entry &
$\texttt{k\_beta\_alpha\_rdot}$ &
Neutral, unused interface entry for this study. \\
\bottomrule
\end{tabular}
\vspace*{-1mm}
\end{table*}


\subsection{Trajectory Optimization and MPC Details}
\label{app:optimizer_mpc_details}

The long-horizon optimizer uses scaled variables $\hat{x}=D_x^{-1}x$ and $\hat{u}=D_u^{-1}u$ and enforces an implicit spatial defect,
\begin{equation}
\hat{x}_{k+1}-\hat{x}_{k}
=
\frac{\Delta s_k}{\Sigma}
F_s(\hat{x}_{k+1},\hat{u}_k,\kappa_k,\theta_k),
\qquad
k=0,\ldots,N_{\mathrm{TO}}-2,
\label{eq:implicit_spatial_defect_app}
\end{equation}
where $\Sigma$ is the spatial horizon length and $F_s$ denotes the normalized version of $dx/ds$. The compact inequality constraint $g_{\mathrm{TO}}\le0$ in Sec.~\ref{subsec:live_trajopt_mpc} collects normalized state bounds, input-rate bounds, lateral track constraints, and the engine-torque envelope
$T_{\min}(\omega_r)\le T_{\mathrm{eng}}\le T_{\max}(\omega_r)$.

The MPC is not minimum-time; it tracks the latest trajectory-optimization reference using a short-horizon quadratic objective,
\begin{equation}
\begin{aligned}
\min_{x_{0:N_{\mathrm{MPC}}},u_{0:N_{\mathrm{MPC}}-1}}
\quad
&\sum_{k=0}^{N_{\mathrm{MPC}}-1}
\left(
\|y_k-y_k^{\mathrm{ref}}\|_{Q}^{2}
+\|u_k-u_k^{\mathrm{ref}}\|_{R}^{2}
+\|u_k-u_{k-1}\|_{R_{\Delta}}^{2}
\right)\\
\mathrm{s.t.}\quad
&x_{k+1}=\Phi_{\Delta t}(x_k,u_k,\theta_k),\qquad
g_{\mathrm{MPC}}(x_k,u_k)\le0 .
\end{aligned}
\label{eq:mpc_tracking_app}
\end{equation}
Here $y_k$ is the tracked output vector and $\Phi_{\Delta t}$ is the discrete-time MPC prediction model. The reported MPC solve time in Fig. \ref{fig:appendix_full_state_comparison} is the wall-clock time required to solve one receding-horizon update during the online closed-loop run. Table~\ref{tab:online_stack_app} summarizes the asynchronous rates and roles of the three online modules used in the closed-loop experiment. The estimator runs at the lowest rate to update preview tire parameters, while the trajectory optimizer replans at an intermediate rate and the MPC tracks the latest reference at the highest rate.

\subsection{Online Data Flow and Additional Results}
\label{app:online_dataflow_results}

At runtime, the MPC tracks the current reference and reports the vehicle progress $s$. The estimator queries visual observations at preview points along the latest reference and returns posterior means for the local tune vectors. These vectors form $P_\theta$, which is used by the next trajectory-optimization solve to generate an updated surface-aware reference. The resulting reference is then streamed back to the MPC.

\begin{table}[!h]
\centering
\caption{Roles of the online modules in the closed-loop experiment.}
\label{tab:online_stack_app}
\footnotesize
\setlength{\tabcolsep}{4pt}
\renewcommand{\arraystretch}{1.05}
\begin{tabular}{lll}
\toprule
\textbf{Module} & \textbf{Role} & \textbf{Relative rate} \\
\midrule
Vision-Estimator & Infer preview tune vectors $\theta_k$ & Lowest ($0.5~Hz$) \\
Trajectory optimizer & Generate surface-aware reference & Intermediate ($2-5~Hz$)\\
MPC & Track latest reference in closed loop & Highest ($20-50~Hz$)\\
\bottomrule
\end{tabular}
\vspace*{-1mm}
\end{table}


\begin{figure*}[!htbp]
\centering
\includegraphics[width=1\textwidth]{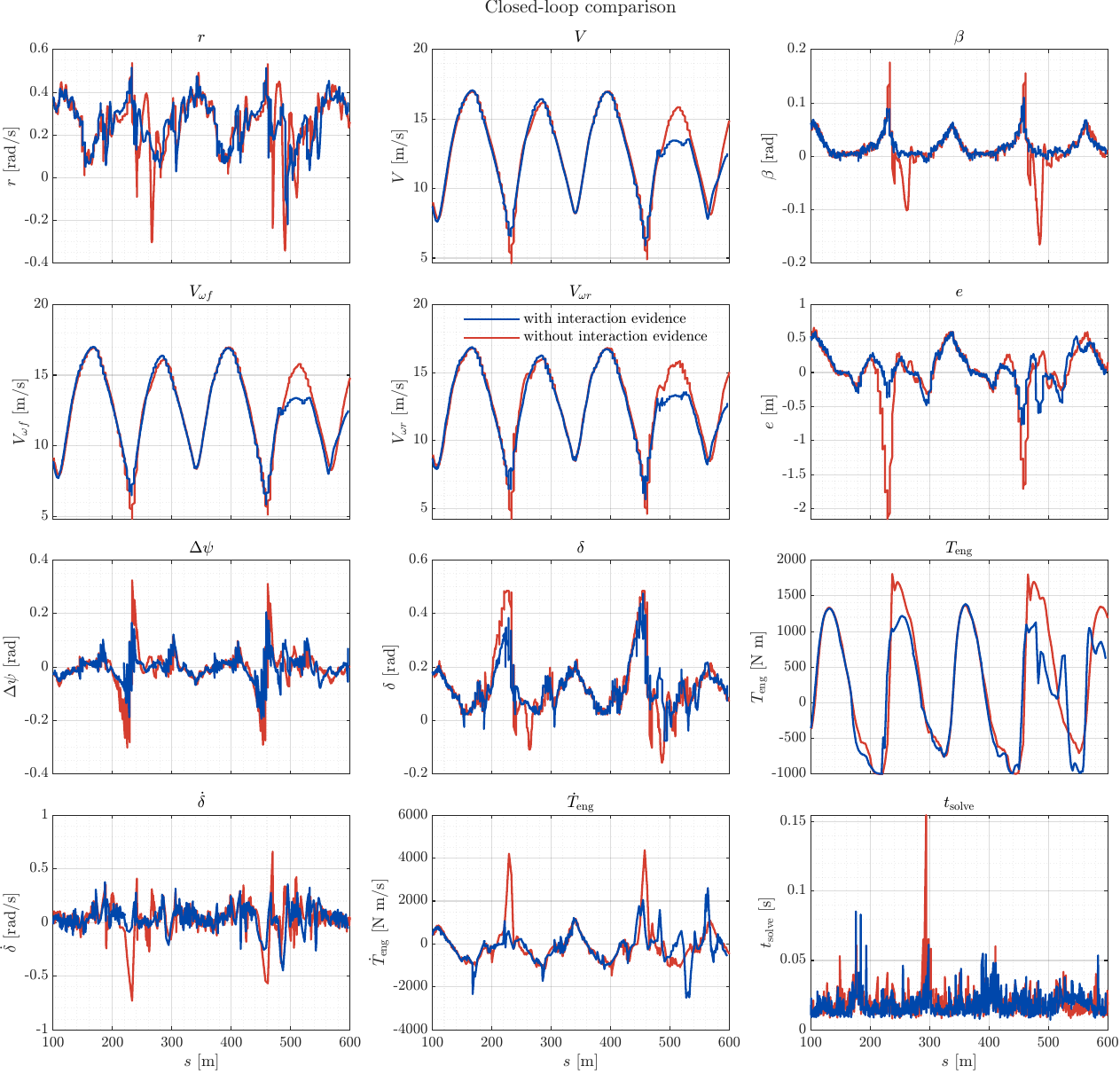}
\vspace*{-2mm}
\caption{Full closed-loop comparison over multiple low-friction encounters. The panels show vehicle states, inputs, and MPC wall-clock solve time for the vision-informed controller and the baseline without online physical-parameter adaptation. Videos of the run can be found at \url{https://tinyurl.com/vision-param-estimation}}
\label{fig:appendix_full_state_comparison}
\vspace*{-2mm}
\end{figure*}

Figure~\ref{fig:appendix_full_state_comparison} reports the full closed-loop state, input, and MPC solve-time comparison over multiple low-friction encounters. On high-friction portions of the track, the controller with interaction evidence and the baseline without interaction evidence exhibit similar speed, wheel-speed, and yaw-rate profiles, indicating that both controllers track comparable references when the nominal tire model is adequate. Differences become pronounced near the low-$\mu$ segments: without interaction evidence, the controller repeatedly enters the low-friction region too aggressively, producing larger spikes in sideslip, lateral error, heading error, steering, and engine-torque rate. In contrast, the interaction-informed controller updates the tire parameters online, resulting in a more conservative surface-aware reference and reduced tracking transients through the low-friction region. The solve-time trace reports the wall-clock time required for each receding-horizon MPC update; the controller remains real-time over the closed-loop run, with only isolated solve-time spikes during difficult transient regions.

\subsection{Generative prior}
\label{app:controls:prior}

The conditional prior $p_0(\cdot\mid e_q)$ used in the racing experiment is a flow model over the $21$-dimensional tire-parameter vector conditioned on the embedding $e_q$. The prior is trained with flow matching on synthetically constructed embedding--parameter pairs. Embeddings assigned to ice receive noisy targets centered on the ground-truth concrete tire-parameter vector $\theta_{\mathrm{concrete}}$; embeddings assigned to snow receive noisy targets centered on $0.9\theta_{\mathrm{concrete}}+0.1\theta_{\mathrm{snow}}$, where $\theta_{\mathrm{snow}}$ is the ground-truth snow vector. These deliberately biased targets provide an imperfect prior for evaluating correction through interaction evidence; the simulated surface parameters retain their ground-truth values.

The trunk has two hidden layers of width $256$ with $64$-dimensional time embeddings; we train with AdamW at learning rate $3\times10^{-4}$, weight EMA (decay $0.999$), for $\sim\!12$k iterations, and use a single integration step at deployment.

\subsection{Support-aware handoff during a drive}
\label{app:controls:handoff}

\begin{figure}[!hbt]
    \centering
    \includegraphics[width=\linewidth]{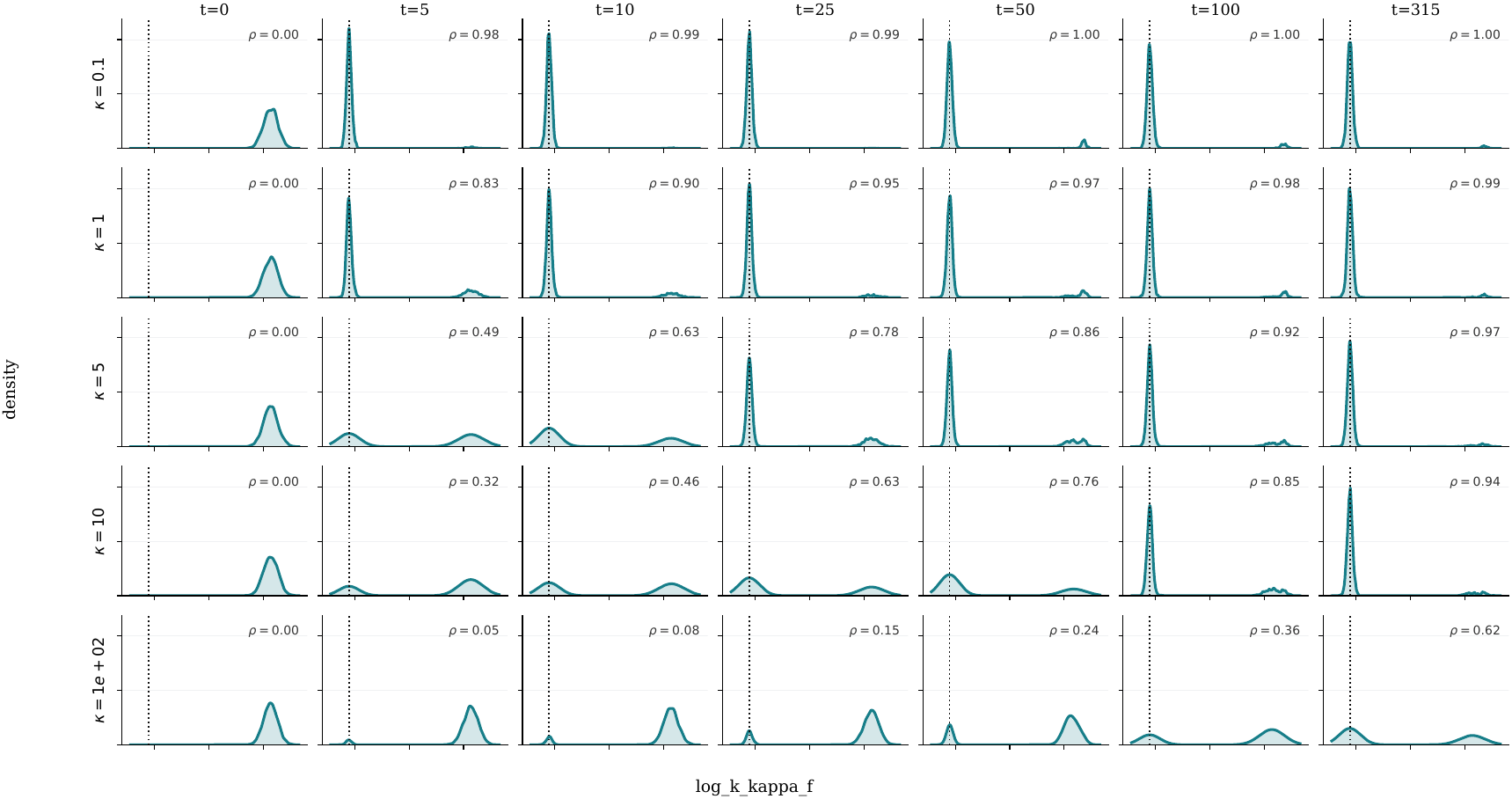}
    \caption{Support-aware handoff from prior to history. Posterior marginal for one tire dimension as the interaction history grows (left to right, $t$) for several support cutoffs (top to bottom, $\kappa$), with the induced $\rho$ annotated per panel and the true surface value marked (dotted). The posterior begins at the prior ($\rho=0$) and sharpens onto the correct value as evidence accumulates ($\rho\to1$); smaller $\kappa$ makes the handoff faster.}
    \label{fig:support_fusion_handoff}
\end{figure}

Fig.~\ref{fig:support_fusion_handoff} illustrates the support-aware fusion of Sec.~\ref{sec:method} on the racing setup. Each panel shows the posterior marginal for a representative tire dimension; columns increase the amount of history $t$, and rows vary the support cutoff $\kappa$. We sweep $\kappa$ and report the induced $\rho=M/(M+\kappa)$. With no history ($t=0$) the posterior is the prior; as interaction evidence accumulates, $M$ grows, $\rho\!\to\!1$, and the posterior expectation converges onto the true surface value (dotted line). 

\clearpage
\raggedbottom

\section{Real-vehicle parameter prediction}
\label{app:hardware}

\noindent\begin{minipage}[t]{0.57\textwidth}
\vspace{0pt}
\raggedright
\setlength{\parskip}{6pt}
The vehicle in Figure~\ref{fig:hardware_vehicle} is equipped with automated steering, throttle, braking, and perception. We infer local Gaussian parameter beliefs by trajectory-based nonlinear log-likelihood inversion using measured states and controls. Trajectory fits provide the means; a temperature-scaled Gauss--Newton inverse-Hessian approximation provides the covariances.

For the grass-driving evaluation, retrieval uses a cosine-similarity cutoff of $0.90$, kernel scale $10$, $\kappa=2$, and $\alpha=0$. Concrete embedding--measurement pairs from a separate run serve as distractors. Section~\ref{sec:hardware} reports the recorded-control prediction results.
\end{minipage}\hfill
\begin{minipage}[t]{0.39\textwidth}
\vspace{0pt}
\centering
\includegraphics[width=\linewidth]{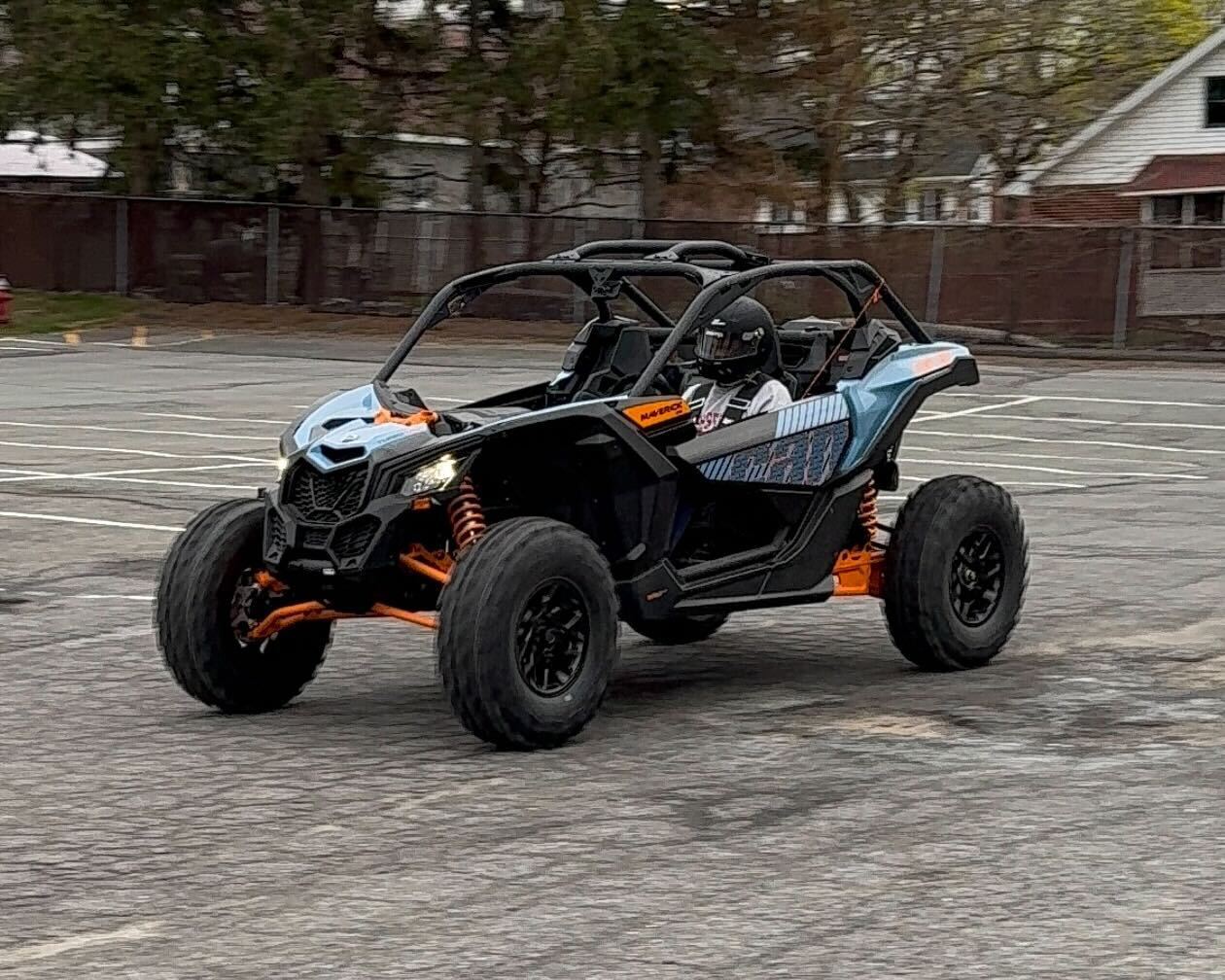}
\captionsetup{hypcap=false}
\captionof{figure}{Autonomous off-road vehicle used in Section~\ref{sec:hardware}.}
\label{fig:hardware_vehicle}
\end{minipage}
\section{Effect of Multimodality on Collision-Risk Estimation}
\label{app:multimodal_risk}

The importance of preserving multimodality becomes apparent when evaluating collision risk under uncertain physical parameters. Consider a surface that may correspond to either high-friction concrete or low-friction black ice, yielding a bimodal friction distribution. We compare this bimodal belief with a moment-matched Gaussian having the same mean and variance.

\begin{figure}[H]
\centering
\vspace{-15pt}
\includegraphics[
    width=.95\textwidth,
    trim=0 0 0 25,
    clip
]{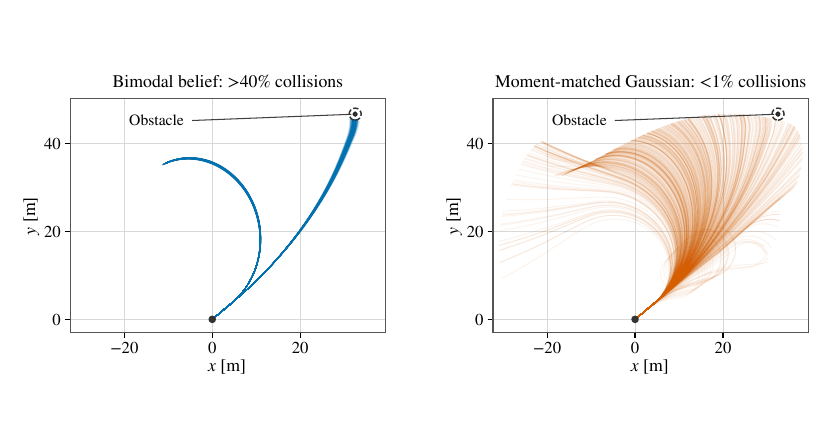}
\vspace{-14pt}
\caption{Predicted vehicle trajectories under a bimodal friction belief (left) and its moment-matched Gaussian approximation (right). Both use the same initial state, fixed control sequence, and obstacle. Despite matching the first two moments, the Gaussian approximation substantially underestimates the collision risk associated with the bimodal belief.}
\label{fig:multimodal_risk}
\end{figure}

To evaluate the effect of these different belief representations, we sample friction parameters from each distribution and propagate the resulting trajectories through the nonlinear vehicle dynamics described in Appendix~\ref{app:vehicle_dynamics}. All rollouts begin from the same initial state and follow the same fixed control sequence near an obstacle. Figure~\ref{fig:multimodal_risk} shows the resulting trajectories.

Despite having identical means and variances, the two beliefs produce substantially different collision rates. More than $40\%$ of rollouts sampled from the bimodal belief collide with the obstacle, compared with fewer than $1\%$ of those sampled from the moment-matched Gaussian. This illustrates how collapsing distinct physical modes into a single Gaussian can substantially underestimate collision risk. While this experiment isolates the effect of belief representation under fixed controls, it motivates risk-aware controllers that explicitly reason over the full mixture posterior rather than relying on posterior summaries.

